\pdfoutput=1
\def\year{2016}\relax
\documentclass[letterpaper]{article}
\usepackage{import}
\usepackage{enumitem}
\usepackage{tabularx}
\usepackage{xspace}
\usepackage{color}
\usepackage{soul}
\usepackage{graphicx}
\usepackage{url}
\usepackage{aaai16}
\usepackage{times}
\usepackage{helvet}
\usepackage{courier}

\newcommand{\eg}{\textit{e.g.},\xspace}
\newcommand{\ie}{\textit{i.e.},\xspace}
\newcommand{\etc}{\textit{etc.}\xspace}

\begin{document}
%
\title{Inferring Fine-grained Details on User Activities and Home Location from Social Media: Detecting Drinking-While-Tweeting Patterns in Communities}
\author{
Nabil Hossain, Tianran Hu,\\
{\Large \bf Roghayeh Feizi}\\
Dept.~Computer Science\\
University of Rochester\\
Rochester, New York\\
\{nhossain,thu\}@cs.rochester.edu
\And
Ann Marie White\\
Dept.~Psychiatry\\
University of Rochester\\
School of Medicine \& Dentistry\\
Rochester, New York\\
AnnMarie\_White@urmc.rochester.edu
\And
Jiebo Luo,  Henry Kautz\\
Dept.~Computer Science\\
University of Rochester\\
Rochester, New York\\
\{jluo,kautz\}@cs.rochester.edu
}

\nocopyright

\maketitle

\begin{abstract}
Nearly all previous work on geo-locating latent states and activities from social media confounds general discussions about activities, self-reports of users participating in those activities at times in the past or future, and self-reports made at the immediate time and place the activity occurs. Activities, such as alcohol consumption, may occur at different places and types of places, and it is important not only to detect the local regions where these activities occur, but also to analyze the degree of participation in them by local residents.  
In this paper, we develop new machine learning based methods for fine-grained localization of activities and home locations from Twitter data. We apply these methods to discover and compare alcohol consumption patterns in a large urban area, New York City, and a more suburban and rural area, Monroe County.
We find positive correlations between the rate of alcohol consumption reported among a community's Twitter users and the density of alcohol outlets, demonstrating that the degree of correlation varies significantly between urban and suburban areas.
While our experiments are focused on alcohol use, our methods for locating homes and distinguishing temporally-specific self-reports are applicable to a broad range of behaviors and latent states.
\end{abstract}

\section{Introduction}
\label{sec:introduction}

Analysis of Twitter has become a widespread approach for geo-spatial studies of human behavior, such as alcohol consumption~\cite{kershaw2014towards,culotta2013lightweight} and exercise~\cite{young2010twitter}, and human latent states, such as sickness~\cite{paul2011you,Modeling_spread_of,sadilek2013nemesis} and depression~\cite{dos2015using,nambisan2015social,tsugawa2015recognizing}.  
However, nearly all prior work, with the notable exception of~\cite{lamb2013separating}, does not attempt to distinguish mere mentions of activities or states from self-reports of activity. Moreover, no attempt has been made to distinguish reports about future or past activities and in-the-moment reports that provide finer details 
when geo-tagged tweets are used to map specific locations of activities. Further insights into the geo-location of activities can be obtained by inferring the home locations of the subjects involved.  
Home location helps analyze the number of members of a community engaging in an activity, the kinds of places where the activity occurs (\eg home, commercial establishment, public place, \etc), and the distance people travel from home to participate in it.
Prior research has used simple heuristics for predicting a social media user's home location, such as the place from which the user most frequently tweets, or the most common last location of the day for the user's posts~\cite{Beware_of_What,We_Know_Where,Friendship_and_Mobility,Socio_spatial_Properties}. But such heuristics are inaccurate for a large percentage of users, \eg in cases when users frequently visit multiple places.

We apply machine learning techniques on Twitter content to identify in-the-moment reports of user behaviors and to accurately predict users' home locations within 100 meters. Using these tools, we develop new methods for a task of critical interest in public health: discovering patterns of alcohol use in urban and suburban settings. Such methods can help us better understand the occurrence, frequency, and settings of alcohol consumption, a health-risk behavior, and can lead to actionable information in prevention and public health. 

Excessive alcohol use has a tremendous negative impact on health and communities. Drinking directly results in about 75,000 deaths annually in the US, making it the nation's third leading cause of preventable death~\cite{centers2004alcohol}. 
Previous research~\cite{kuntsche2005young,naimi2003binge} 
shows that social factors play an important role in developing drinking patterns over time. While social media such as Twitter is both ubiquitous and publicly available, little research has investigated the relationship between virtual social contexts and the alcohol referencing or alcohol-linked behaviors found there
in various real-world community settings. 

In this paper, we aim to predict where Twitter users are when they report on drinking. We report on several stages of work to accomplish this research objective. First, we collected geo-tagged tweets from urban, suburban, and rural areas of New York State. Using human computation, we created a training set that captures important details such as whether the tweet mentions drinking alcohol, the user drinking, or the user drinking at the time of tweeting. We created a hierarchy of three support vector machine (SVM) classifiers~\cite{burges1998tutorial} to distinguish tweets up to these fine details. 
Each of these SVMs achieves an F-score above $83\%$ and is used to classify tweets from New York City and from Monroe County, a predominantly suburban area in upstate New York containing one medium-sized city (Rochester), in order to develop methods that can perform in ``big city'' as well as ``small city'' contexts of social media use.

We also performed fine-grained home location inference of Twitter users to generate community descriptions, such as to calculate the proportion of ``social media drinkers'' drinking at home, and to analyze how far people travel from home to drink-and-tweet. Existing home inference methods either rely on continuous and expensive GPS data, covering a small number of users, or suffer from poor accuracy. We trained an SVM classifier to predict home location for \emph{active users} (users with as little as 5 geo-tagged tweets) within 100 by 100 meter grids. Considering the sparse and noisy nature of Twitter data that poses serious challenges in pinpointing where people live, our classifier achieves a high accuracy of 70\%, covering 71\% active users in New York City. We also investigated ways to balance granularity and coverage. Prior work on home location has been limited to localizing at the city level; ours is the first to achieve block-level accuracy.

\section{Related Work}
\label{sec:related-work}
\subsection{Latent States \& Activities from Social Media}
Most prior work on using Twitter data about users' online behavior
has estimated aggregate disease trends in a large geographic area, typically at the level of a state or a large city. Researchers have examined influenza tracking
\cite{culotta2010towards,achrekar2012twitter,Modeling_the_Impact,dredze2013flu,brennan2013towards},
mental health and depression
\cite{golder2011diurnal,de2013predicting}, as well as general public
health across a broad range of diseases~\cite{brownstein2009digitaldisease,paul2011you}.  
Some researchers have begun modeling health and contagion across individuals~\cite{ugander2012structural,horvitz_health_individual,de2013predicting}.
For example,~\cite{sadilek2012predicting} showed that Twitter
users who exhibit symptoms of influenza can be accurately detected
using a language model based on word trigrams.  
A detailed epidemiological model can be subsequently
built by following the interactions between sick and healthy
individuals in a population, where physical encounters
are estimated by spatio-temporal collocated tweets.
nEmesis~\cite{sadilek2013nemesis} scored restaurants
in New York City by the number of Twitter users who posted status updates from 
a restaurant and within the next several days posted self-reports of symptoms of food poisoning.
Our hierarchical classifiers use the same kind of word-trigram features at each level.

Little prior work has attempted to distinguish true in-the-moment self-reports on Twitter from more general discussion of a condition or activity. A notable exception is~\cite{lamb2013separating}, which explored language models that could distinguish
discussion of the flu from self-reports. This work enriched the set of n-gram language features by including manually-specified sets of words, features for hashtags and retweets, and various syntactic patterns. For separating general discussion from reports of some particular person being sick, n-grams were most important, followed by the manually-specified word classes. For separating reports of the user being sick from reports of others being sick, n-grams were again most important, by the hashtag/retweet features. The overall success of n-grams supports our n-gram based approach for latent activity detection. The authors did not use hierarchical classifiers or attempt to distinguish in-the-moment-reports from those about the past or future.

\subsection{Alcohol Consumption}
Despite the huge public health costs exacted by alcohol use, commercial interests and individuals, for example, teens~\cite{moreno2009display,egan2011alcohol} do post about alcohol and drinking in social media.
Alcohol-related posts are seen as credible reports by teens and thus posts can influence perceived social norms, a factor linked to the uptake of drinking behaviors~\cite{litt2011adolescent}. 

In the case of alcohol use, social context certainly matters.  For instance, survey research shows that having close friends that drink heightens alcohol use and perceptions about alcohol use in teen life, in general~\cite{jackson2014predictors,polonec2006evaluating}. 
Peer alcohol consumption behavior of one's social network, particularly those of relatives and friends (not immediate neighbors and co-workers), is a risk factor for alcohol use, specially among adolescents~\cite{rosenquist2010spread,ali2010social}.

When the geography of one's daily life creates proximity to alcohol (\ie greater spatial/temporal availability of on-premise or off-premise alcohol outlets, \etc), a well-documented risk factor for alcohol use and its array of related adverse public health consequences emerges~\cite{campbell2009effectiveness,weitzman2003relationship,holmes2014impact,scribner1999alcohol,scribner2008contextual,livingston2008alcohol,livingston2008longitudinal,livingston2011longitudinal,kypri2008alcohol,chen2010community,scribner1994alcohol,zhu2004alcohol,britt2005neighborhood,liang2011revealing}.
Modifying proximity is often explored as a public health policy means to reduce alcohol use, for instance, in neighborhoods~\cite{sparks2011regulating}. However, the association between neighborhood alcohol outlet density and percentage of alcohol consumers may be more complex due to variation in travel patterns and neighborhood styles, and mediated by proximity to one's home (\eg within one-mile)~\cite{schonlau2008alcohol}.

Social media is a new ubiquitous source of real-time community and individual public-health related behaviors. When seeking to apply social media to detect the social media ecology of health behaviors such as alcohol use, it is important to identify not only whether but where (the settings in which) the mentions or posts are occurring.  As both geo-physical and virtual access to rapidly diffused messages about alcohol and its use may heighten risky drinking and related behaviors, methods are needed to permit the study of these potentially interacting influences. 
Such methods can reveal different risk patterns associated with different locations not prior known, and help inform more localized or targeted intervention strategy development. For instance, as social network structures are observable in social media,  and as ``neighbor'' attributes can influence drinking behavior among online friends or followers, studying network influence in social media settings like Twitter may illuminate drinking risk patterns not previously known.  

However, current methods for examining these influences are very limited. Methods for detecting problematic alcohol use in communities are typically opportunity or survey based (e.g., driver check-points, community surveys, ED admissions, or health care-based screenings), not often scalable to population levels due to resource restrictions. Research on how to vivify a community's raft of social media posts to detect its alcohol use patterns is only now starting to emerge. For instance,~\cite{Tamersoy:2015:CSD:2700171.2791247} distinguished long-term versus short-term drinking/smoking abstinence from the social media site Reddit. These researchers were able to use linguistic features from content posted, and social interaction features derived from users' network structure through the application of supervised learning.
In this paper, we propose new automated methods for identifying both whether and where self-reports of drinking are occurring among Twitter users in two major metropolitan regions of New York State.

\subsection{Home Location Detection}
With the knowledge of home locations, we can gain a better insight to human mobility patterns, as well as lifestyle in general. In~\cite{Socio_spatial_Properties,Friendship_and_Mobility,Exploiting_Place_Features}, home location is the key origin to calculate the distance that people travel and to estimate the distance between social network users in a pairwise fashion. Home location has also been used to model individuals' living conditions and lifestyles~\cite{Modeling_the_Impact}.  We organize the discussion of related work on home location prediction by the type of data used.

\subsubsection{Language content}
There has been much prior work on using language features in non-geotagged social media posts to predict the home locations of users at a coarse grain, at the level of a city or state. In~\cite{Where_is_This}, linguistic features and place names from tweets were used to create a classifier that infers home locations at city, state and time zone levels in the top 100 most populated US cities with accuracies of 58\%, 66\%, and 78\% respectively. This suggests that language models are not good for fined-grained home localization (in our case, within 100 meters). Similar results, accurate at most to several kilometers, appear in~\cite{Beware_of_What}. In~\cite{You_Are_Where}, the authors used a content-based method to detect Twitter users' home cities, placing $51\%$ of active users within 100 miles of their actual home locations. 

\subsubsection{Geo-tagged Data}
Others developed ``single-attribute'' models based on different social network features, for example, taking the value of users' ``Employment'' as their home cities in Google+, or using geo-tags in FourSquare, Google+, and Twitter posts to predict the city.
Geo-tagged Foursquare data was used in~\cite{We_Know_Where} to infer home cities within 50 kilometers with 78\% of user coverage. A dataset containing the traces of 2 million mobile phone users from a European country was used in~\cite{Friendship_and_Mobility} to estimate home locations based on the places with most check-ins. The paper reported that by manual checking, the most check-ins method achieved 85\% accuracy when the area was divided into 25 by 25 km cells. 

Other researchers used simple heuristics to select the home location from the set of locations in a user's geo-tagged posts. The most popular heuristics are to assume that the location with the most check-ins is home~\cite{Socio_spatial_Properties}, or to assume that the common last location of the day from which one tweets is home~\cite{Modeling_the_Impact}. The accuracy and coverage of such heuristic approaches was not reported. We discovered that these prior methods individually suffered from low accuracy and/or coverage. For example, the most check-ins approach performs poorly when a user visits several places with similar frequencies. 

\subsubsection{Wearable GPS and Diary Data}
GPS and diary data make home detection more precise and easier because they are more dense and continuous than social network location data, but they are more expensive to obtain, resulting in low population coverage when used in locating homes. In~\cite{Inference_Attacks_on}, a device recorded location coordinates every several seconds when the car was moving on 172 subjects' vehicles. The subjects reported the ground truth of their homes. The authors then used 4 heuristic algorithms to compute the coordinates of each subject's home, and found that the best one was ``last destination of a day''. 
The median distance error of their best algorithm was 60.7 meters. In~\cite{Enhancing_Security_and}, the researchers performed agglomerative clustering on the GPS traces of users until the clusters reached an average size of 100 meters. Next they manually eliminated clusters with no recorded points between 4PM and midnight and those falling outside the residential areas.

Semantically labeling places is another important topic related to home location detection. In~\cite{Far_Out_Predicting}, the authors used GPS data from 307 people and 396 vehicles, then divided the world into 400 by 400 meter grids, and assigned each GPS reading to the nearest cell. They found that the top 10 frequently visited locations can usually be semantically labeled as ``home'', ``work'', ``favorite restaurant'' and so on. Other researchers~\cite{Placer_Semantic_Place} performed experiments using two diary datasets --- American Time Use Survey and the Puget Sound Regional Council Household Activity Survey --- where each location had a semantic label such as ``home'' or ``school''. They extracted several features of a location and trained place classifiers using machine learning, reporting a classification accuracy above 90\% on locations labeled as ``home''.

\section{Alcohol Usage Detection}
\label{sec:alcohol-prediction}
\begin{figure}
\centering
\includegraphics[width=0.9\columnwidth]{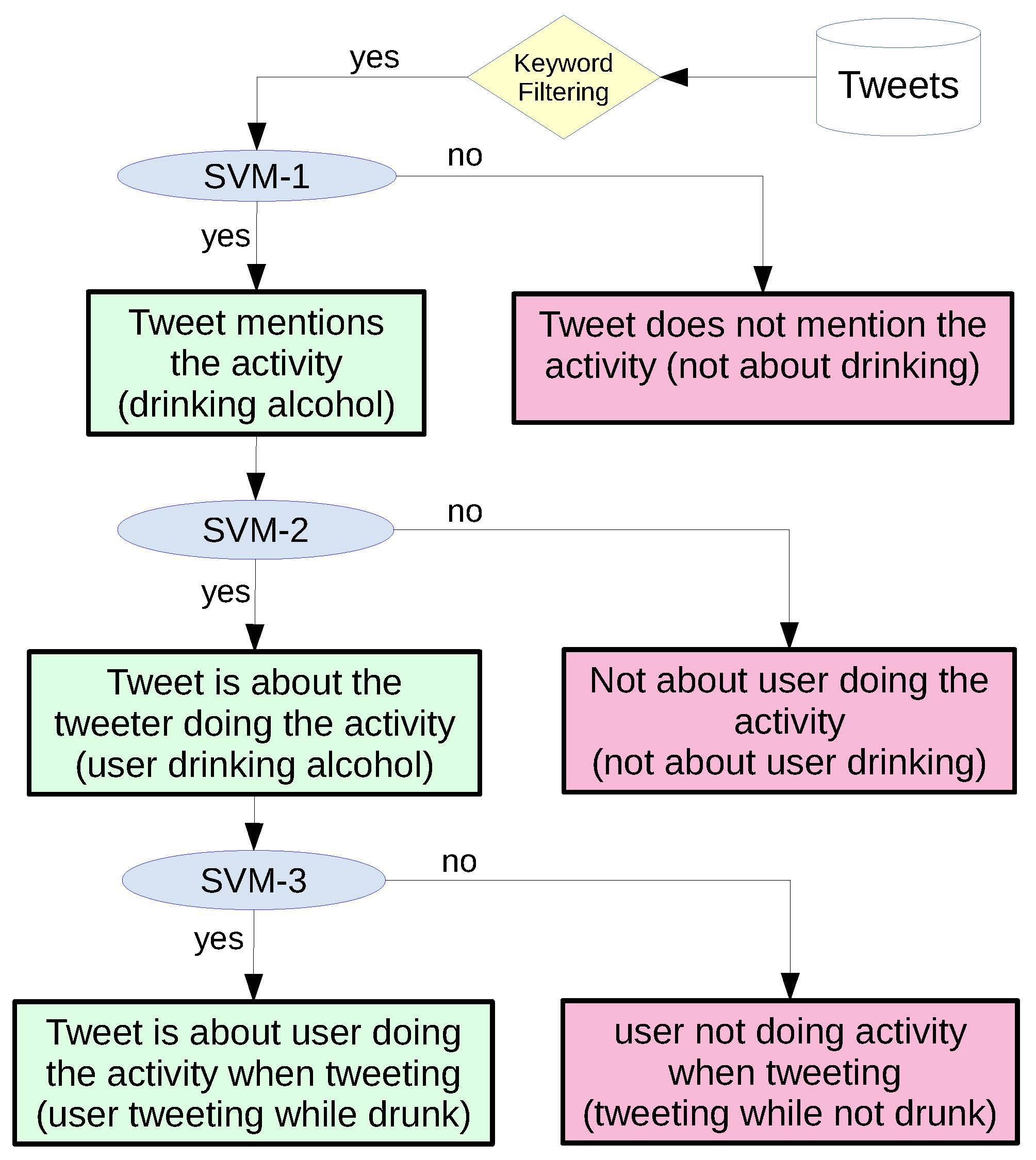}
\caption{Flowchart for latent activity detection.}
\label{fig:hierarchical-flowchart}
\end{figure}
We now describe our methods for detecting geo-temporal alcohol consumption via Twitter. We discuss the data preparation steps, the hierarchical classification approach, the strategies we employed to reduce classifier overfitting and the results.

\subsection{Ground Truth}
We collected geo-tagged tweets from urban, suburban and rural areas in New York State from July 2013 to July 2014. Similar to the approach used in~\cite{paul2011you}, we began the process of creating a training dataset by first filtering tweets if they included a mention of alcohol, defined by the inclusion of any one of several drinking-related keywords (\eg ``drunk'', ``beer'', ``party'') and their variants. The word set was reviewed and modified with local community member input from our social media analytic advisory group, the Big Data Docents. 

We were interested in labeling each tweet that passed this filter by applying a hierarchy of three yes/no feature questions, as follows:

\begin{description}
\item[Q1:] Does the tweet make any reference to drinking alcoholic beverages?
\item[Q2:] if so, is the tweet about the tweeter him or herself drinking alcoholic beverages?
\item[Q3:] if so, is it likely that the tweet was sent at the time and place the tweeter was drinking alcoholic beverages?
\end{description}

We labeled this {\bf Alcohol dataset}\footnote{dataset and keywords available in: \url{cs.rochester.edu/u/nhossain/icwsm-16-data.zip}} using the Amazon Mechanical Turk\footnote{\url{http://www.mturk.com}}. Given a tweet, a turker was asked Q1, and only if the turker answered ``yes'', then he/she was asked Q2, and so on. Each question was passed to three Turkers and the answer choices were ``yes'', ``no''and ``not sure''.
Tweets that didn't receive consensus in turker ratings ( (yes/no) agreement among less than two turkers) were discarded from the dataset. The remaining tweets were labeled `1' if two or more turkers answered ``yes'', otherwise they were labeled `0' for each feature question. 
Since for each tweet the questions were asked hierarchically, the approach resulted in a smaller ground truth for deeper questions, as Table~\ref{tab:alcohol-dataset} shows.

\begin{table}
\centering
\begin{tabular}{l||c|c|c|} 
                    &       Q1      &       Q2      &       Q3  \\ \hline \hline
Class size (0, 1)   & 2321, 3238    & 579, 2044     & 642, 934  \\ \hline
Precision           &       0.922   &   0.844       &   0.820   \\ \hline
Recall              &       0.897   &   0.966       &   0.845   \\ \hline
F-score             &       0.909   &   0.901       &   0.833   \\ \hline
\end{tabular}
\caption{Alcohol dataset test results}
\label{tab:alcohol-dataset}
\end{table}

\subsection{Dataset Pre-processing}
Tweet texts are usually conversational texts, noisy and unstructured, making it difficult to create a good feature set using them. We performed several pre-processing techniques to reduce lexical variation in tweets. 
These include converting hyperlinks to ``\texttt{$\#$url}'', mentions to ``\texttt{$\#$mention}'', emoticons to positive and negative emoticon features, 
using hashtags as distinct features, and truncating three or more consecutive occurrences of a character in a word to two consecutive occurrences (e.g. ``druuuuuuunk'' $\rightarrow$ ``druunk''). Using the pre-processed tweets and their labels, we created separate trigram linguistic feature sets for the three questions. In order to reduce overfitting, we only kept the top $N$ most-frequent features, where $N = 25\%$ of the size of the training set size for the corresponding question.

\begin{table}
\centering
\begin{tabular}{rl||rl} 
neg. features  & weights & pos. features  & weights \\ \hline \hline
club	&	-1.244	&	drunk	&	1.056	\\
shot	&	-1.206	&	beer	&	1.028	\\
party	&	-1.193	&	wine	&	0.998	\\
\texttt{$\#$}turnup	&	-0.972	&	alcohol	&	0.936	\\
yak	&	-0.919	&	vodka	&	0.9	\\
lean	&	-0.919	&	drink	&	0.899	\\
crown	&	-0.823	&	tequila	&	0.857	\\
root beer	&	-0.772	&	hangover	&	0.854	\\
root	&	-0.772	&	drinking	&	0.811	\\
wasted	&	-0.745	&	liquor	&	0.793	\\
turn up	&	-0.673	&	\texttt{$\#$}beer	&	0.779	\\
turnup	&	-0.668	&	hammered	&	0.757	\\
binge	&	-0.663	&	take shot	&	0.749	\\
drunk in love	&	-0.593	&	alcoholic	&	0.749	\\
in love	&	-0.52	&	get wasted	&	0.715	\\
water	&	-0.501	&	champagne	&	0.708	\\
turnt up	&	-0.499	&	booze	&	0.692	\\
fucked up	&	-0.441	&	ciroc	&	0.68	\\
fucked	&	-0.441	&	rum	&	0.653	\\
water bottles	&	-0.423	&	whiskey	&	0.635	\\
\end{tabular}
\caption{Top weighted features for SVM-1}
\label{tab:topSVM-1-feats}
\end{table}

\subsection{Training}
For each of the three questions, we trained a linear support vector machine (SVM) to predict the answer. As shown in Figure~\ref{fig:hierarchical-flowchart}, these SVMs are hierarchical~\cite{koller1997hierarchically}. For example, the data for SVM-2 (SVM for question Q2) include only the tweets labeled by SVM-1 as ``yes'' and for which consensus was reached by turkers for Q2. This restricts the dataset distribution as we go down the hierarchy. Compared to a single flattened multi-class classifier, hierarchical classifiers are easier to optimize, and because they have a restricted feature set, we can build more complex models without overfitting. This way of classifying tweets is also more intuitive and suits our purposes. In other words, SVM-1 will be specialized to filter drinking-related tweets, while SVM-3 assumes that the input tweet is about drinking and particularly the tweeter drinking, and decides whether the tweeter was drinking when he/she posted the tweet. 

For each SVM, we used $80\%$ of the labeled data for training and the remaining $20\%$ for testing. We applied 5 fold cross validation to reduce overfitting and used the F-score for model selection. The F-score, ranging between 0 and 1, is the harmonic mean of precision and recall, and the higher the score the lower the classification error.

\begin{table}
\centering
\begin{tabular}{rl||rl} 
neg. features  & weights & pos. features  & weights \\ \hline \hline
she	&	-1.222	&	will	&	0.411	\\
he	&	-0.936	&	when you	&	0.37	\\
your	&	-0.87	&	bad	&	0.358	\\
people	&	-0.841	&	when drunk	&	0.334	\\
they	&	-0.676	&	with	&	0.318	\\
are	&	-0.658	&	am	&	0.303	\\
my mom	&	-0.623	&	get drunk	&	0.301	\\
drunk people	&	-0.6	&	through	&	0.3	\\
guy	&	-0.551	&	drink	&	0.296	\\
\texttt{$\#$}mention you	&	-0.5	&	dad	&	0.292	\\
her	&	-0.472	&	us	&	0.286	\\
for me	&	-0.454	&	friday	&	0.283	\\
baby	&	-0.447	&	more	&	0.282	\\
their	&	-0.431	&	still	&	0.28	\\
his	&	-0.423	&	little	&	0.28	\\
see	&	-0.417	&	drinking	&	0.28	\\
most	&	-0.394	&	free	&	0.27	\\
talking	&	-0.377	&	pong	&	0.263	\\
the drunk	&	-0.368	&	already	&	0.261	\\
\end{tabular}
\caption{Top weighted features for SVM-2}
\label{tab:topSVM-2-feats}
\end{table}

\subsection{Results}
The results in Table~\ref{tab:alcohol-dataset} show high precision and recall for each question. They also suggest that the more detailed the question becomes, the harder it gets for the classifier to predict correctly. 
This is not unexpected because intuitively we expect Q3 to be a harder question to answer compared to Q1. More importantly, our hierarchical classification approach shrinks the training data as we go down to deeper questions, most likely making it difficult for the classifiers down the hierarchy to learn from the smaller data. However, we believe that this approach is better than a multi-class SVM approach which, although would use the full training data to answer each question, does not have the advantage of restricting the data distribution to focus on the question. For example, Table~\ref{tab:topSVM-1-feats} shows that SVM-1 mainly uses features related to alcoholic drinks to determine whether the tweet is related to drinking alcoholic beverages. SVM-2 distinguishes self-reports of drinking from general drinking discussion by using pronouns and implicit references to drinking, as Table~\ref{tab:topSVM-2-feats} suggests. Table~\ref{tab:topSVM-3-feats} shows that, having known that the tweet is related to the user drinking alcohol, SVM-3 identifies drinking in-the-moment using temporal features (\eg ``hangover'', ``last night'', ``now'') and features related to the urge to drink (\eg ``need'', ``want'').  

\begin{table}
\centering
\begin{tabular}{rl|rl} 
neg. features  & weights & pos. features  & weights \\ \hline \hline
hangover	&	-1.179	&	\texttt{$\#$}url	&	0.662	\\ 
need	&	-1.088	&	shot	&	0.461	\\ 
want	&	-0.878	&	here	&	0.429	\\ 
was	&	-0.67	&	\texttt{$\#$}mention when	&	0.4	\\ 
when	&	-0.617	&	bottle of wine	&	0.387	\\ 
or	&	-0.605	&	drank	&	0.368	\\ 
real	&	-0.601	&	now	&	0.36	\\ 
alcoholic	&	-0.6	&	think	&	0.352	\\ 
for	&	-0.561	&	one	&	0.349	\\ 
last night	&	-0.525	&	good	&	0.327	\\ 
will	&	-0.525	&	vodka	&	0.318	\\ 
wanna	&	-0.523	&	by	&	0.312	\\ 
tonight	&	-0.52	&	me and	&	0.312	\\ 
got	&	-0.492	&	outside	&	0.307	\\ 
weekend	&	-0.483	&	hammered	&	0.304	\\ 
yesterday	&	-0.471	&	haha	&	0.3	\\ 
was drunk	&	-0.47	&	drive	&	0.3	\\ 
\end{tabular}
\caption{Top weighted features for SVM-3}
\label{tab:topSVM-3-feats}
\end{table}

\section{Home Location Prediction}
\label{sec:home-prediction}
Existing home inference methods suffer from either low coverage (GPS \& diary data) or coarse granularity and low accuracy (language models and prior work on geo-tagged data), making them inadequate for problems that require both high coverage and fine granularity. Our more sophisticated machine learning based algorithm 
combines a number of different features describing each user's daily trajectories as determined from geo-tagged tweets, thus predicting users' home locations from sparse tweets with high accuracy and coverage. We now describe our method for home location prediction of Twitter users, the creation of a labeled training data, the feature set, our results, and we evaluate our system.

\subsection{Dataset \& Pre-Processing}
We collected geo-tagged tweets sent from the greater New
York City area during July 2012 and from the Bay Area during
06/01/2013 - 08/31/2013. A typical geo-tagged tweet contains the ID of the poster, the exact coordinates from where the tweet was sent, time stamp, and the text content.
Due to the inherent noise in the geo-tags, we split the areas into
100 by 100 meter grids and treat the center of each grid as the target of home detection. Each tweet is assigned to its closest grid, and every time a user's tweet appears in a grid we say the user has a {\bf check-in} in this grid.
Similar to previous work~\cite{limits,refined,predictability}, we only focus on users who have sent at least 5 geo-tagged tweets, and we call them {\bf active users}. Also following these studies, we take each user's {\bf hourly traces} (only one location for each hour in our sampling duration) instead of using every single check-in. Thus, if a user appears in several unique grids in an hour, we take the grid with
the highest number of check-ins as the user's location for the hour (ties are broken by random selection). If a user's location is not observed in an hour, the location for that hour is set to ``Null''.
Typically, the hourly traces $T_{U}$ of a user $U$ form a sparse vector, for example, $T_U = [Null, L_{i}, Null, ..., L_{j}]$, and the size of $T_U$ is the number of hours in the sampling period. We provide a snapshot of our dataset in Table~\ref{tab:data_statistics}.
\begin{table}
  \centering
  \resizebox{\columnwidth}{!}{
  \begin{tabular}{|c|c|c|c|}
    \hline
    & NYC & Bay Area \\
    \hline
    No. of tweets & 2,636,437 & 3,633,712 \\
    \hline
    Total no. of active users & 55,237 & 53,314 \\
    \hline
    No. of tweets annotated by AMT & 5,000 & 5,000\\
    \hline
    No. of ground-truth homes & 1,063 & 987\\
    \hline
  \end{tabular}
  }
  \caption{Description of our dataset for home inference.}
  \label{tab:data_statistics}
\end{table}

\subsection{Ground Truth}
Obtaining fine-grained ground truth is challenging because it involves identifying a Twitter user's home from several locations the user checked-in without being told by the user. Some researchers relied on information from user profiles~\cite{Beware_of_What,We_Know_Where,Where_is_This}, others manually inspected the detection results~\cite{Friendship_and_Mobility}.
However, the location information in user profiles is coarse (at city level), while manual inspection is not scalable.
Reading a tweet that says ``Enjoying the beautiful conference room view!'', a human can tell that it was sent from a workplace. Tweets such as ``finally home!'' or ``home sweet home'' 
are most likely sent from home. Thus, we relied on tweet content and human intelligence to build the ground truth for home location.

We asked faithful Twitter users what they would like to post when at home. Based on their answers, we selected a set of 50 keywords (\eg ``home'', ``bath'', ``sofa'', ``TV'', ``sleep'', \etc) and their variants which are likely to be mentioned in tweets sent from home. Next, we filtered tweets that contained at least one of these keywords. Then, we relied on Amazon Mechanical Turk to find the tweets sent from home. Each turker was given a questionnaire containing 5 tweets to answer. For each tweet we asked: ``is this tweet sent from home?'', and the options were ``yes'', ``no'' and ``not sure''. Each questionnaire was answered by three unique turkers. We only retained the tweets which, all three turkers believed, were sent from home.

\subsection{Features Based on Human Mobility}
Previous work using linguistic features from tweet content~\cite{Where_is_This,You_Are_Where} did not achieve good accuracy in granular settings, and even in course-grained conditions these methods required over a few hundred tweets per user to obtain reasonable accuracy. Our goal is to predict homes for users with as little as 5 tweets to increase coverage. Therefore, instead of using linguistic features, we extract features that capture temporal and spatial properties of homes. Although some of these features alone
(e.g. check-in frequency, PageRank score) 
can be used as reasonable baseline methods to detect homes, we show that combining features appropriately using a machine learning method brings significant gain in both accuracy and coverage. 
We now discuss how we obtain these features from a user's hourly traces and how they capture inherent properties of home. 

\subsubsection{Check-in Rate}
As we discussed earlier, taking the location of most check-ins as home is a popular method. Throughout the paper, we refer to this method and the corresponding feature as ``Most Check-in''.
Although check-in based methods for home detection work well on GPS data~\cite{Inference_Attacks_on}, they perform much worse on Twitter data. This is because GPS devices keep recording locations every few seconds whereas the frequency of a user's geo-tagged tweets are low and largely vary based on the type of user. The location with most check-ins definitely is important to a user, but that does not necessarily mean it is the home. 

\begin{figure}[!b]
\centering
\includegraphics[width=1\columnwidth]{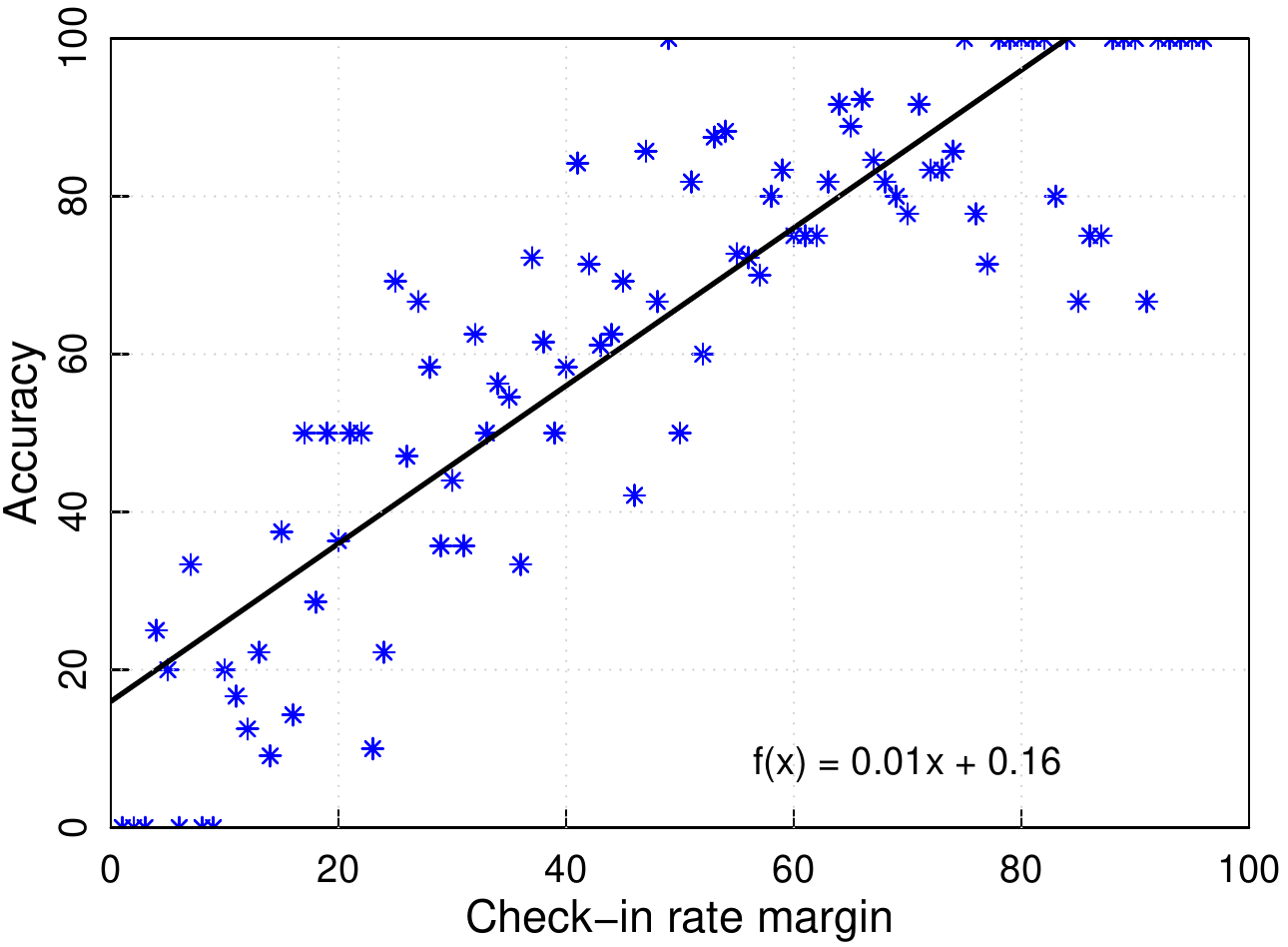}
\caption{Accuracy of home inference using Most Check-in feature vs. margin between the locations with most check-ins and second most check-ins. For each point ($X$, $Y$), $X$ is the margin and $Y$ is the home detection accuracy obtained using Most Check-in feature for the group of users in our ground truth having margin $X$.
}
\label{fig:checkin_limit}
\end{figure}

\begin{figure}[!ht]
\centering
\includegraphics[width=8cm,height=6cm,keepaspectratio]{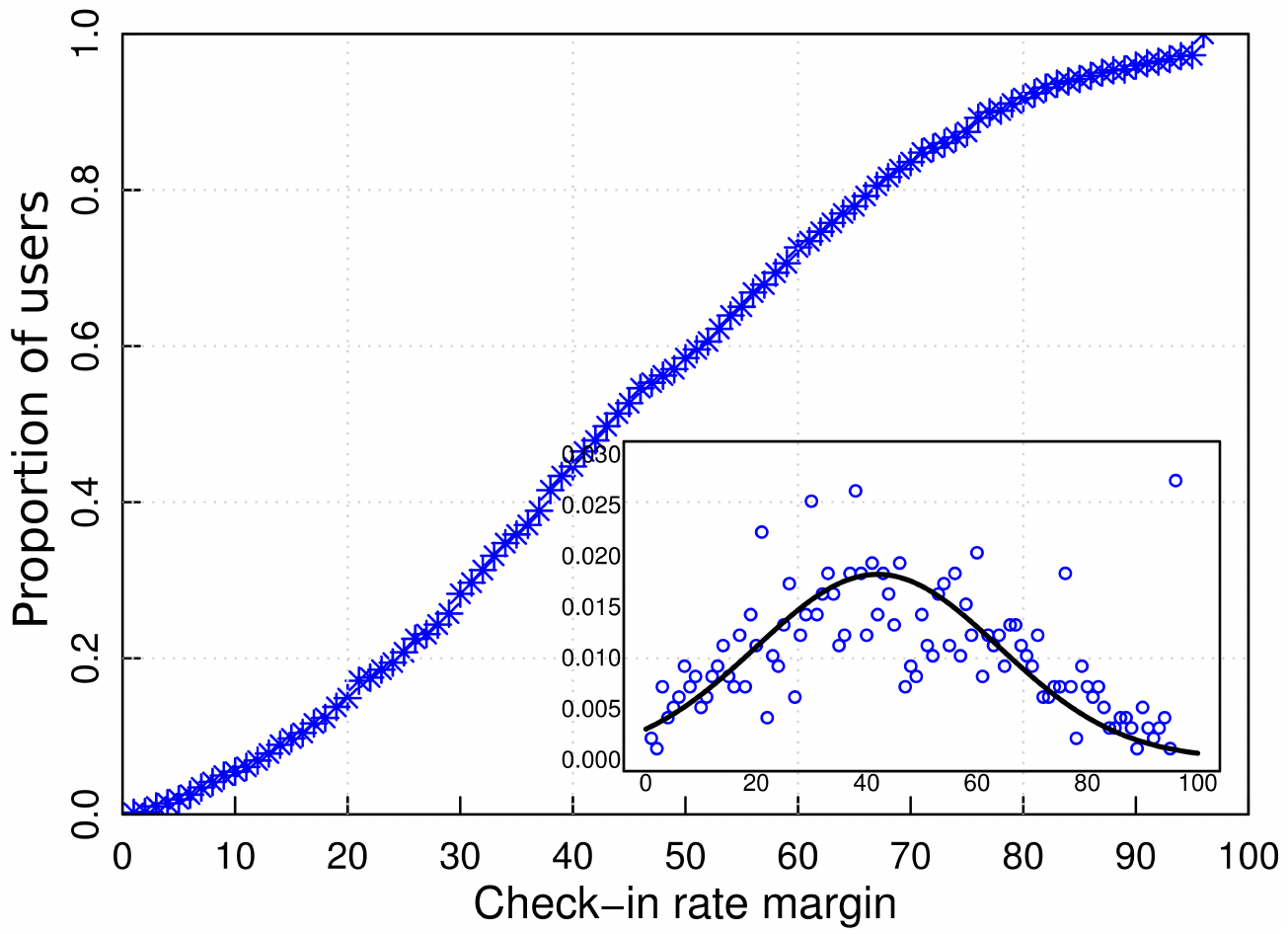}
\caption{The cumulative distribution of the margin between the locations with most check-ins and second most check-ins. The smaller plot shows the distribution of margin between most check-in and second most check-in locations.}
\label{fig:checkin_cover}
\end{figure}

For user $U$, we define the {\bf margin} between two locations of check-ins $A$ and $B$ as $P_A - P_B$, where $P_A$ and $P_B$ are percentages of $U$'s check-ins at $A$ and $B$ respectively. Figure~\ref{fig:checkin_limit} shows that for a user, the lower the margin between the most check-in location and the second most check-in location, the less effective is the Most Check-in feature as an accurate predictor of home. For instance, the accuracy is 70\% only when this margin is 50 or higher.
Figure~\ref{fig:checkin_cover} shows that only a small number of users have large margins between most check-in and second most check-in locations (\eg only about 20\% of the users have margins above 70, which means that home detection accuracy for these users using Most Check-in method is about 80\%, according to Figure~\ref{fig:checkin_limit}).
These explain why the Most Check-in method performs poorly in fine-grained settings --- for example, as the grid with most check-in shrinks from 1 by 1 kilometer block to many 100 by 100 meter grids, the most check-in percentage spreads over many of these smaller grids, lowering the margin between the new most check-in location and the new second most check-in location.
To circumvent this problem, we extract 3 features for each location $L$ checked-in by a Twitter user $U$:
\begin{itemize}
\item the percentage of check-ins of $U$ at location $L$
\item the margin between $L$ and those of its immediate higher and lower most check-in locations
\end{itemize}

\subsubsection{Check-in Frequency During Late Night}
Intuitively, the places people check-in at late night are probably their homes. For example,~\cite{Modeling_the_Impact} estimated a person's home by taking the mean of a two-dimensional Gaussian fitted to the person's check-ins between 1AM and 6AM. This method potentially alleviates the biases caused by other frequently visited places during daytime. Thus, for each location visited by a user, we take the check-in percentage of that location computed over a restricted time period of 12AM - 7AM as a feature, which we define as the {\bf late night feature} of that location.

\subsubsection{Last Destination of a Day}
According to research using GPS data~\cite{Inference_Attacks_on}, the last destination of a person on a day (no later than 3AM) is most likely the home, highlighting that people's daily movements end at their homes. Based on this assumption we extract a mobility feature, which we call the {\bf last destination feature}. For each location visited by a user, we count the number of times the location had been the last destination of the day, and we add this count to our feature set. 

\subsubsection{Last Destination with Inactive Late Night}
Since ``last destination'' might suffer from check-ins sent from non-home places (\eg when the night was spent outside), we add to our feature set a variant of last destination. We only consider tweets sent on the days when people were inactive during late night (12AM - 7AM). We exclude the days with active late night and, for each place visited by the user, we count the number of times the place had been the last destination in the remaining days.

The original check-in feature has limitations in obtaining a broader coverage in detecting homes. The above three features introduce extra human behaviour information to the simple check-in feature and help reduce this limitation.

\subsubsection{Temporal Features}
According to~\cite{Inference_Attacks_on}, the probability of being at home varies over time. For each place checked-in by a user, we compute the check-in percentages in that place at each hour of the day over the sampling period, and we add these 24 values (which sum to 100\%) to our feature set. These time related features help us capture the property of home in terms of temporal patterns.

\subsubsection{Spatial Features}
Home is a crucial start/end point of many of our movements. Thus, for each place we add 2 more features --- weighted PageRank~\cite{weighted_pagerank} and Reversed PageRank scores --- to model how importantly a place behaves as an origin and a destination. To apply PageRank, we first transfer a Twitter user's trace into a directed graph called the \emph{movement graph}, in which the vertices are the locations visited by the user and a directed edge from vertex $L_{i}$ to $L_{j}$ represents that location $L_{j}$ is visited directly from $L_{i}$. To quantify the certainty and importance of transitions between locations, we assign a weight to each edge. The weight should be proportional to the number of times a transition appears in the user's trace, and inversely proportional to the number of idle hours during the transition. Thus, assuming that $T$ is the set of hourly traces of a user over the sampling period, the weight $w_{(L_{i}, L_{j})}$ is the ratio of the total number of transitions from $L_i$ to $L_j$ in $T$ to the total number of idle hours during all these transitions.

After constructing a user's movement graph, we apply PageRank to calculate, for each visited location, the importance of that location as a destination.
To study the importance of that location as an origin, we calculate the Reversed PageRank score by reversing each edge direction in the movement graph (edge weights remain unchanged), and then applying weighted PageRank. The PageRank and Reversed PageRank scores describe the spatial characteristics of movements.

\subsection{SVM Training and Home Location Evaluation}
\subsubsection{Training} We trained a linear SVM classifier using all these features to capture important feature combinations that better distinguish homes. Each training datapoint is a tweet identified uniquely by user ID and location ID, labeled ``home'' or ``not home'', having 32 feature values calculated from the user's hourly traces. For each Twitter user, the classifier outputs a score for all the places the user checked-in from. If the place with the highest score exceeds a threshold, it is marked as the user's home. Otherwise, the user's home is marked ``unknown'', which decreases our home detection coverage. Table~\ref{tab:home_SVM_features} shows the most significant SVM features.
\begin{table}
  \centering
  \resizebox{\columnwidth}{!}{
  \begin{tabular}{|c|c|c|}
    \hline
    \textbf{Positive Features} & \textbf{Weight} \\
    \hline
    Check-in ratio & 2.03\\
    \hline
    Margin between top two check-ins& 0.19 \\
    \hline
    PageRank Score & 0.19 \\
    \hline
    Last destination with inactive late night & 0.12 \\
    \hline
    Reversed PageRank score & 0.09 \\
    \hline
    \textbf{Negative Features} & \textbf{Weight} \\
    \hline
    Margin below next higher check-in& -0.30 \\
    \hline
    Margin under next higher PageRank & -0.28 \\
    \hline
    Margin under next higher Reversed PageRank& -0.21 \\
    \hline
    Rank of Reversed PageRank & -0.07 \\
    \hline
    Rank of PageRank & -0.07 \\
    \hline
  \end{tabular}
  }
  \caption{Top SVM features and their weights.}
  \label{tab:home_SVM_features}
\end{table}

\begin{figure}[!ht]
\centering
\resizebox{\columnwidth}{!}{
\begin{tabular}{c}
 \includegraphics[width=\columnwidth]{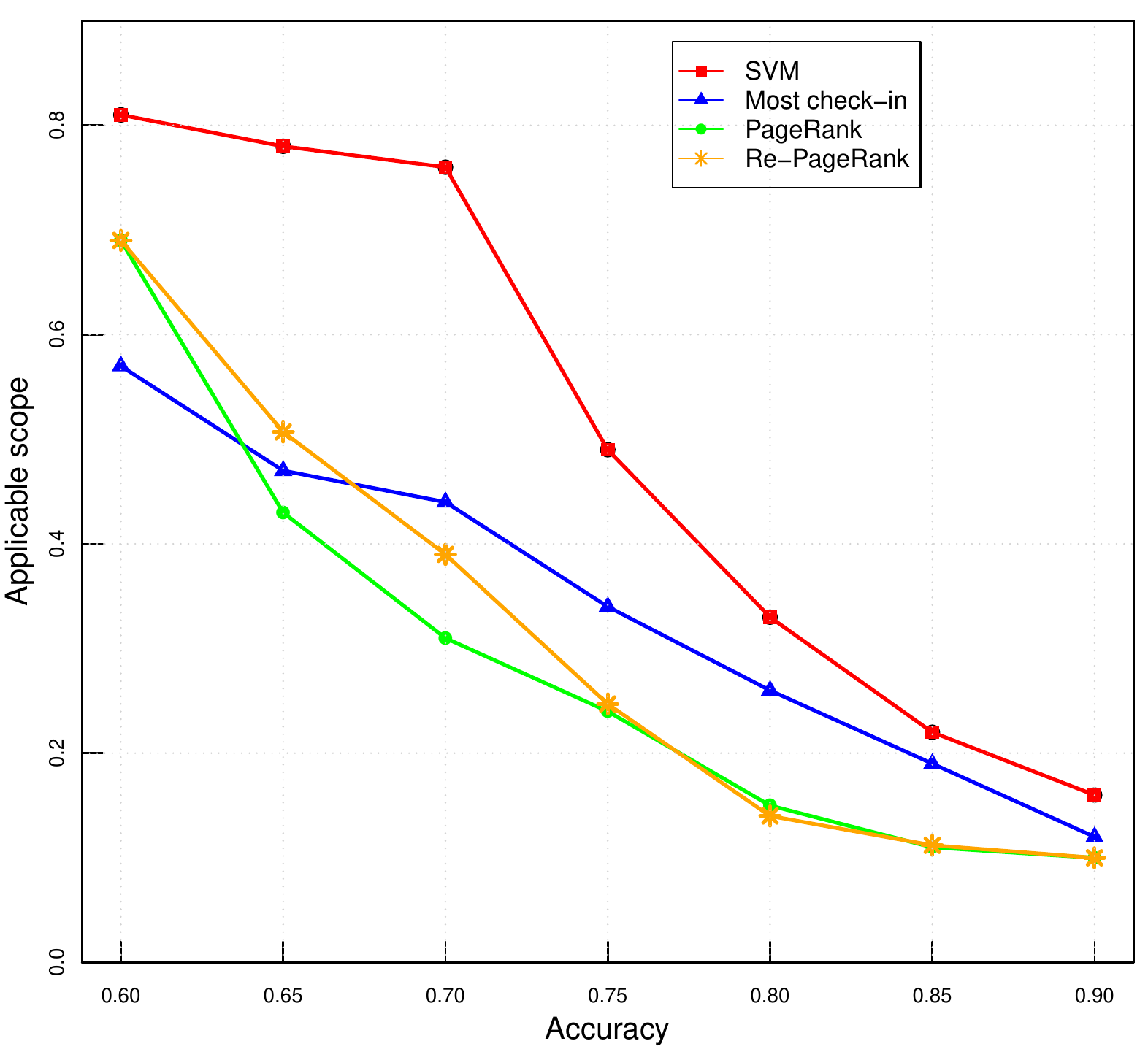}
\end{tabular}
}
\caption{The trade-off between accuracy and coverage for different home detection methods using New York City data.}
\label{fig:coverage_accuracy}
\end{figure}

\subsubsection{Accuracy vs. Coverage}
To guarantee the practicality of our home detection method, we need to balance granularity and coverage. Because of the natural trade-off between granularity and detection accuracy, we fix the granularity to 100 by 100 meter grid and explore the relationship between accuracy and coverage. The accuracy can be adjusted by varying the threshold, which also affects coverage.

Figure~\ref{fig:coverage_accuracy} shows how our methods compare with three other single-feature based methods in terms of accuracy and coverage. The tuning parameter for PageRank (and Reversed PageRank) scores was the extent to which the highest PageRank Score was larger than the second highest one, and for Most Check-ins it was the check-ins count. Homes were not predicted using Most Check-ins when the most check-in count was less than 3. 
At every accuracy level, our method covers more homes than other methods, suggesting that a combined model significantly increases coverage over single-feature based models. 
Particularly, when we set the accuracy of each method to 70\% (which we think is acceptable for urban computing), our classifier obtains 71\% and 76\% coverage in NYC and Bay Area respectively, significantly higher than those achieved using individual features. 
\begin{figure}[!h]
\centering
\includegraphics[width=0.9\columnwidth]{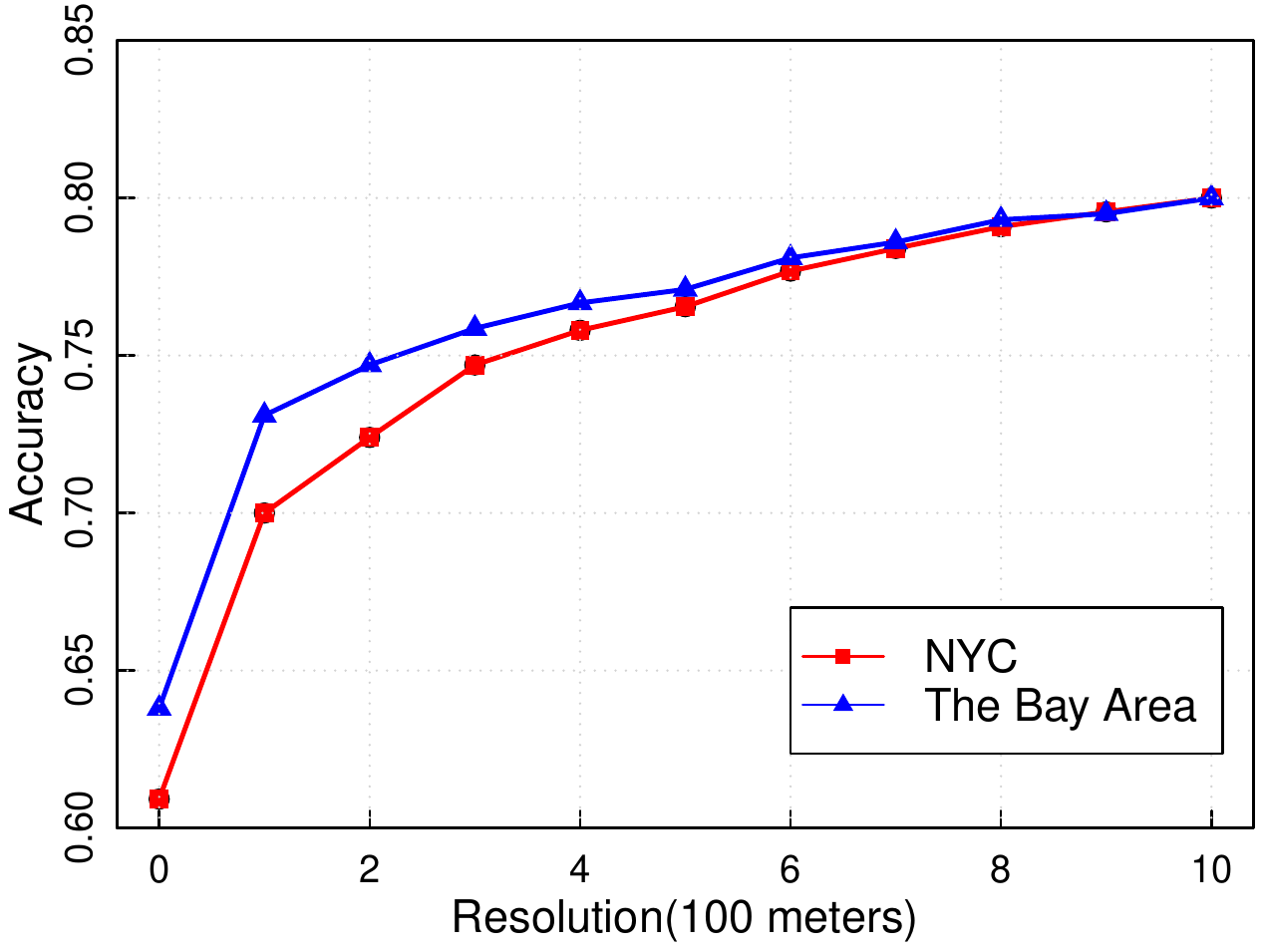}
\caption{Resolution vs. accuracy of home detection.
}
\label{fig:resolution}
\end{figure}

\subsubsection{Granularity}
Since we performed home detection to 100 by 100 meter grids, the {\bf resolution} of this grid-based method is around 70 meters ($\surd2 * 100 / 2 \approx 70$ m). 
We explore how resolution affects our method's accuracy by setting coverage at 80\% and varying the resolution from 100 meters to 1000 meters. Figure~\ref{fig:resolution} shows that increasing the resolution increase the accuracy although the rate of increase of accuracy slows down and peaks at around 80\%. Compared to previous work~\cite{Beware_of_What}, our method provides higher resolution with similar accuracy (~80\%).

\section{Analysis of Alcohol Consumption via Twitter}
\label{sec:discussion}
In this section, we discuss the results obtained by applying our SVMs on geo-tagged tweets from New York City (dataset range: 11/19/2012 - 03/31/2013) and from Monroe County in upstate New York (dataset range: 07/03/2014 - 04/27/2015).
We specifically chose these datasets to study alcohol consumption in urban (NYC) vs suburban (Monroe) settings. 
We analyze drinking at home vs. away from home, and we investigate the relationship between the density of tweets sent from different regions while intoxicated and the density of alcohol outlets in those regions.
The following terms will be used throughout this section:
\begin{itemize}
\item drinking-mention: SVM-1 predicts ``yes''
\item user-drinking: SVM-2 predicts ``yes''
\item user-drinking-now: SVM-3 predicts ``yes''
\end{itemize}

We ran the set of NYC and Monroe tweets in the order shown in Figure~\ref{fig:hierarchical-flowchart}. The results in Table~\ref{tab:alcohol-classification} show that for each drinking-related question, NYC has a higher proportion of tweets marked positive compared to the corresponding proportion in Monroe County. One possible explanation is that a crowded city such as NYC with highly dense alcohol outlets and many people socializing is likely to have a higher rate of drinking happening at a time compared to a suburban area such as Monroe county with low population and alcohol outlet density. 

\begin{table}
\centering
\resizebox{\columnwidth}{!}{
  \begin{tabular}{|l||c|c|}
                                        & NYC       & Monroe        \\ \hline \hline
    No. of geo-tagged tweets            & 1,931,662 &  1,537,979    \\ \hline
    Passed keyword filter               & 51,321    &  26,858        \\ \hline
    drinking-mention                    & 24,258    &  13,108        \\ \hline
    user-drinking                       & 23,110    &  12,178        \\ \hline
    user-drinking-now                   & 18,890    &  8,854         \\ \hline
    Correlation with outlet density     & 0.390     & 0.237         \\ \hline 
  \end{tabular}
  }
  \caption{Classification of drinking-related tweets on NYC and Monroe datasets.}
  \label{tab:alcohol-classification}
\end{table}

\begin{figure}[tb]
\centering
\begin{tabular}{c}
\includegraphics[width=7.2cm,keepaspectratio]{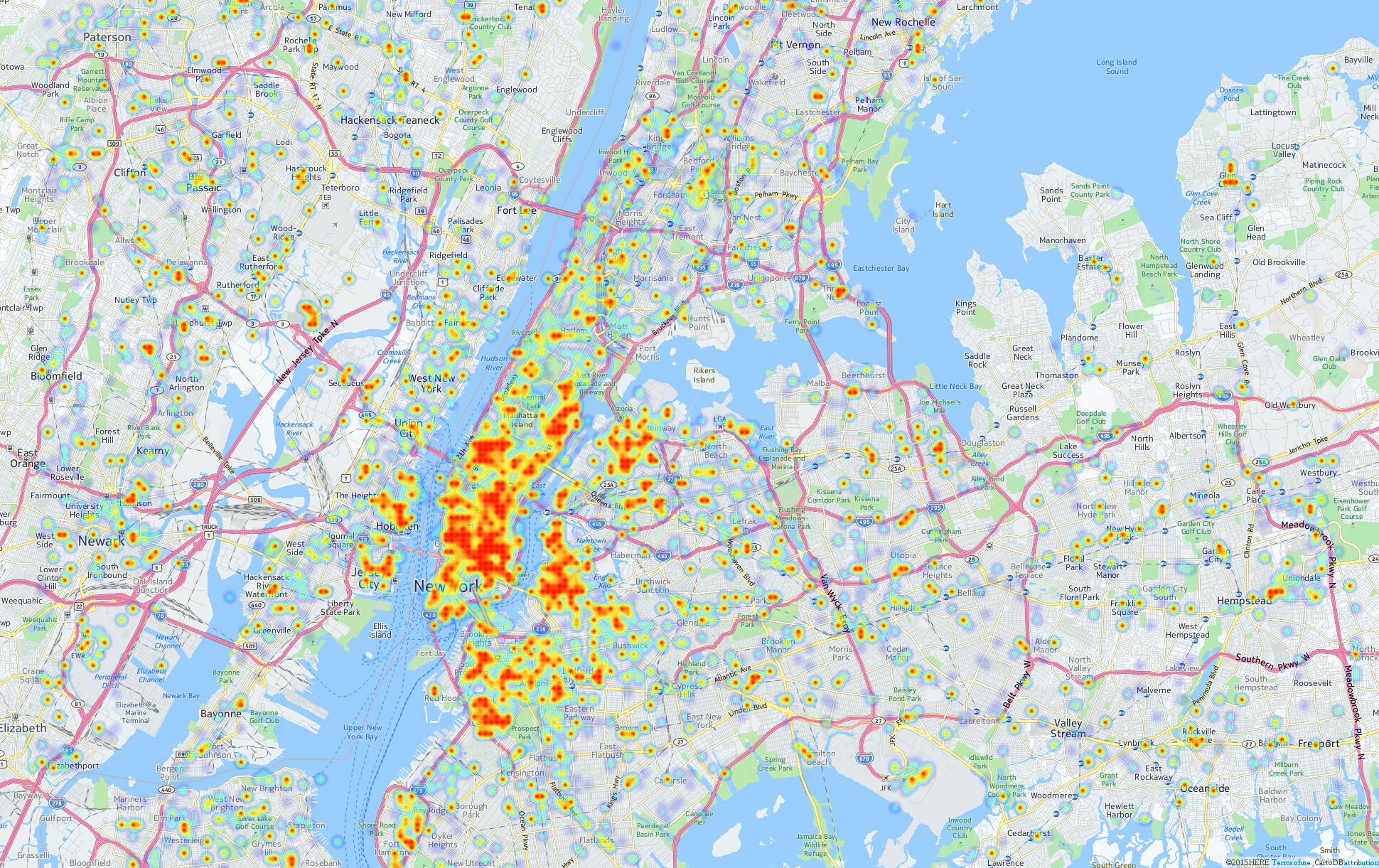} \\
(a) NYC \\ \\
 \includegraphics[width=7.2cm,keepaspectratio]{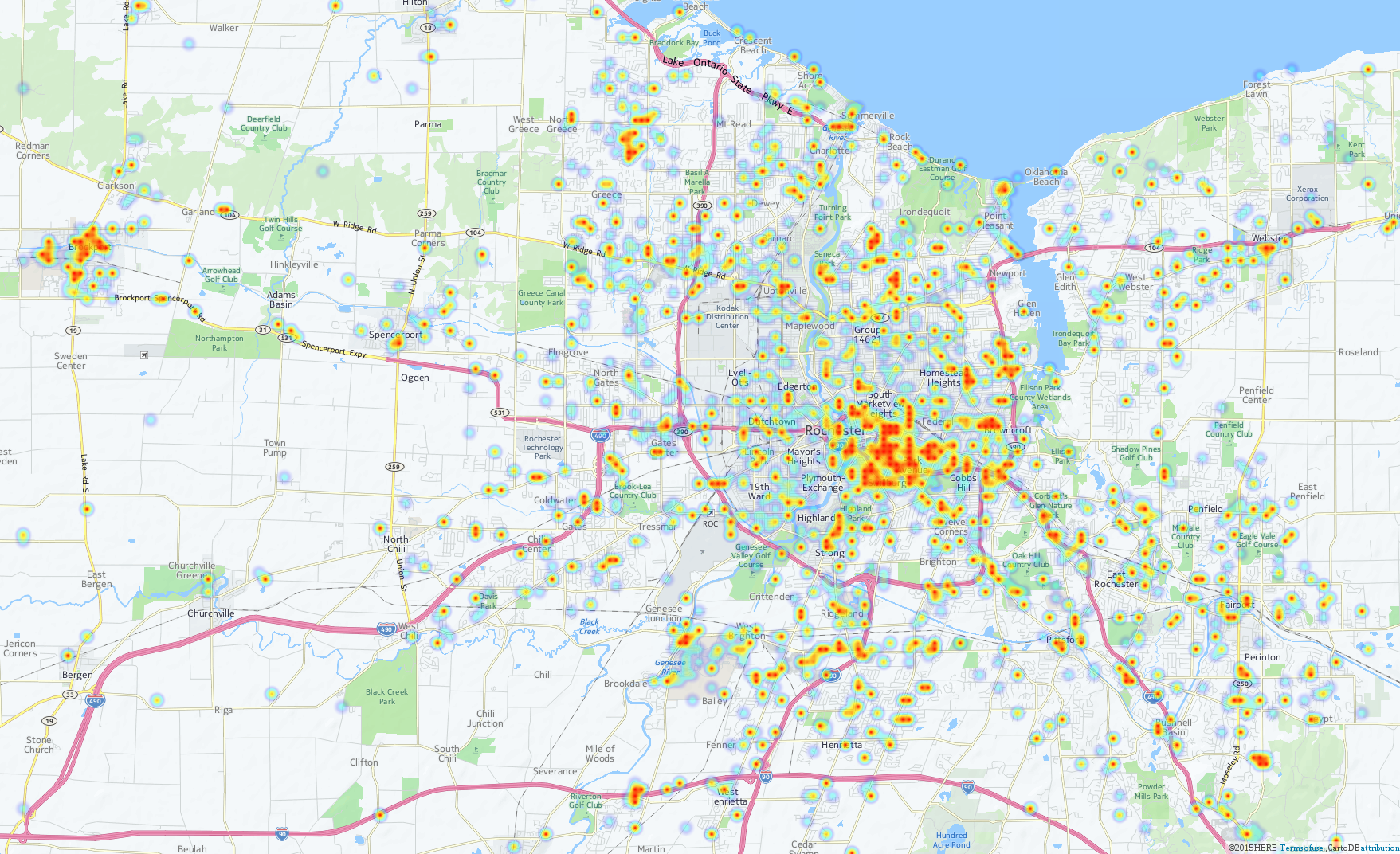} \\
 (b) Monroe
\end{tabular}
\caption{Heat maps of user-drinking-now tweets showing unusual drinking zones. In NYC, the drinking hot spots are Lower Manhattan and it's surroundings whereas in Monroe County they are Downtown Rochester (center) and the city of Brockport (left).}
\label{fig:alcohol-visualization}
\end{figure}

Figure~\ref{fig:alcohol-visualization} shows the zoomed geographic distributions\footnote{obtained using CartoDB --- \url{http://cartodb.com/}} of user-drinking-now tweets via \emph{normalized} heat maps. These maps were constructed by splitting the geographic area for each dataset into 100 by 100 meter grids, then computing the proportion of tweets in each grid that were user-drinking-now (excluding grids that had less than 5 user-drinking-now tweets), and using these values as the degree of ``heat''. That is, the grids with ``more heat'' are those where the proportion of in-the-moment drinking tweets compared to the total geo-tagged tweets are much higher. 
We believe that such grids are regions of unusual drinking activities. 

We also computed the alcohol outlet densities\footnote{obtained from NYS LAMP --- \url{lamp.sla.ny.gov/}} for the grids and then calculated the correlation between the alcohol outlet density and the density of user-drinking-now tweets.
As Table~\ref{tab:alcohol-classification} shows, the density of user-drinking-now tweets in both our datasets exhibit positive correlations with alcohol outlet density, with $p$-values less than $1\%$. Although correlation does not necessarily imply causation, these results agree with several prior work~\cite{campbell2009effectiveness,sparks2011regulating,weitzman2003relationship,scribner2008contextual,kypri2008alcohol,chen2010community} 
which claim that alcohol outlet density influences drinking.

\subsection{Location-based Analysis}
\begin{figure*}
\centering
\begin{tabular}{c}
 \includegraphics[width=0.88\textwidth]{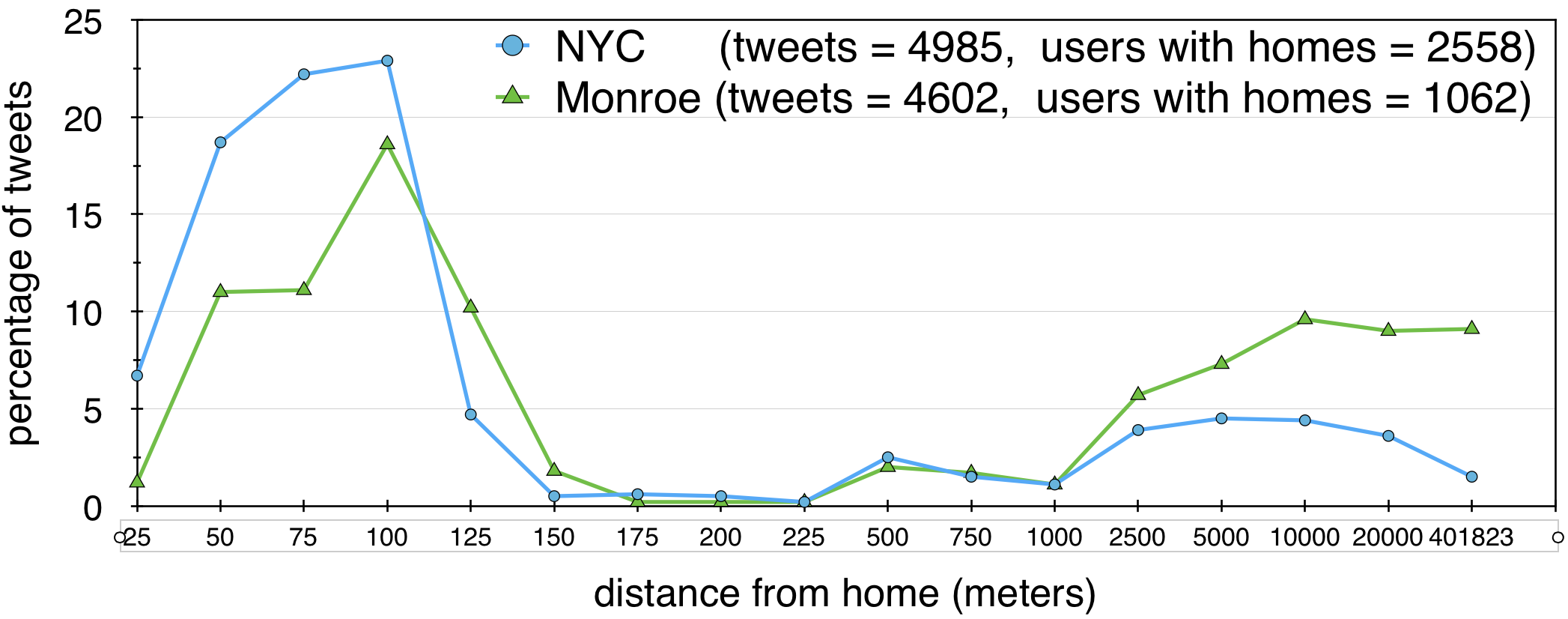}
\end{tabular}
\caption{Histogram of distances from home for tweets sent while the user was drinking.}
\label{fig:drinking-histogram}
\end{figure*}

The ability to detect homes and locations where user-drinking-now tweets are generated enables us to compare drinking going on at home vs. not at home. For this purpose, we only used homes predicted with at least 90\% accuracy which resulted in some loss of coverage (see Figure~\ref{fig:coverage_accuracy}). We filtered all Twitter users with homes in our datasets and extracted all the user-drinking-now tweets posted by these users. For these tweets, we plotted the histogram of distance from home, shown in  Figure~\ref{fig:drinking-histogram}. We see that NYC has a larger proportion of user-drinking-now tweets posted from home (within 100 meters from home) whereas in Monroe County a higher proportion of these tweets generated at driving distance (more than 1000 meters from home).

\section{Discussion and Future Work}
\label{sec:conclusion}
We proposed a machine learning based model for detecting latent activities and user states via Twitter to such fine details that have not been distinguished yet. The model not only distinguishes people discussing an activity vs. discussing themselves performing the activity, but also determines whether they are performing it at-the-moment vs. past/future. We showed the strength of our model by applying it to the detection of alcohol consumption as an example application. Coupled with our other contribution of home location prediction, the model allows us to study Twitter users' drinking behavior from several community or ecological viewpoints built from the fine-grained location information extracted.

Models that permit the fine-grained study of alcohol consumption in social media can reveal important real-time information about users and the influences they have on each other. We can begin to evaluate the merits of these data for public health research. Such analyses can teach us who is and isn't referencing alcohol on Twitter, and in what settings, to evaluate the degree of self-reporting biases, and also help to create a tool for improving a community's health, given social networks can become a resource to spread positive health behaviour. For instance, the peer social network ``Alcoholics Anonymous''\footnote{\url{http://www.aa.org/}} is designed to develop social network connections to encourage abstinence among the members and establish helpful ties. 

Although we apply home localization to describe a geographical community portrait of drinking referencing patterns among its social media users, 
since people spend a large portion of their time at home, our model enables a wide range of applications that were previously impractical. For instance, we can analyze human mobility patterns; we can study the relationship between demographics, neighborhood structure and health conditions in different zip codes, thus understanding many aspects of urban life and environments. Research in these areas and alcohol consumption is mainly based on surveys and census, which are costly and often incur a delay that hamper real-time analysis and response. Our results demonstrate that tweets can provide powerful and fine-grained cues of activities going on in cities. 

While Twitter use is ubiquitous, its users are not a representative sample of the general population; it is known to include more young and minority users~\cite{life_project}. Bias, however, is a problem in any sampling method. For example, surveys under-represent the segment of the population that is unwilling to respond to surveys, such as undocumented immigrants. 
Statistics estimated from Twitter (or any other source) can be adjusted to account for known biases by weighting data appropriately. While addressing Twitter's bias is beyond the scope of this paper, our methods can permit further work in this area by locating users in communities with fine-grained detail, meaning more fine-grained demographic data becomes available for linkage. We also note that the average sampling rate of US Census in each state is about 3\%~\cite{census:2011}, which is similar to the percentage of users we covered out of all the Twitter users.

Our future work will perform a comprehensive study of alcohol consumption in social media around features such as user demographics, settings people go to drink-and-tweet (\eg friends' house, stadium, park), \etc We can explore the social network of drinkers to find out how social interactions and peer pressure in social media influence the tendency to reference drinking. Another interesting study is to compare the rate of in-flow and out-flow of drinkers in adjacent neighborhoods. All these analyses will help us understand the merits of these methods for analyzing drinking behavior, via social media, at a large-scale with very little cost, which can lead to new ways of reducing alcohol consumption, a global public health concern. Finally, our models are broadly applicable to various latent activities and make way for future work in many other domains. 

\section{Acknowledgements}
Research reported in this publication was supported by the National Institute of General Medical Sciences of the National Institutes of Health under award number R01GM108337, the National Science Foundation under Grant No. 1319378 and the Intel ISTCPC. The content is solely the responsibility of the authors and does not necessarily represent the official views of the NIH and the NSF. The authors thank members of the Big Data Docents, our community collaborative research board, for their guidance in this scientific work. 

\small
\bibliographystyle{aaai}
\bibliography{www-2016-geo-drinking,nemesis-bibliography-file,Home_Location}
\end{document}